\documentclass[letterpaper, 10 pt, conference]{ieeeconf}  

\IEEEoverridecommandlockouts                              

\usepackage{cite}
\usepackage{graphicx}
\usepackage{amsmath}
\usepackage{amssymb}
\usepackage{subfigure}
\usepackage{stfloats}
\usepackage{algpseudocode}
\usepackage{url}
\usepackage{tabularx}
\usepackage{array}
\usepackage{multirow}
\usepackage{color}
\usepackage{booktabs}
\usepackage{enumerate}
\usepackage{breqn}
\usepackage{indentfirst}
\usepackage{bm}

\title{\LARGE \bf
	MRPB 1.0: A Unified Benchmark for the Evaluation of Mobile Robot Local Planning Approaches
}

\author{Jian Wen$^{1}$, Xuebo Zhang$^{1\dagger}$, Qingchen Bi$^{2}$, Zhangchao Pan$^{3}$, Yanghe Feng$^{4}$,\\ Jing Yuan$^{1}$, and Yongchun Fang$^{1}$ 
	\thanks{This work is supported in part by National Key Research and Development Project under Grant 2018YFB1307503, in part by Tianjin Science Fund for Distinguished Young Scholars under Grant 19JCJQJC62100, in part by Tianjin Natural Science Foundation under Grant 19JCYBJC18500, and in part by the Fundamental Research Funds for the Central Universities.}
	\thanks{$^{1}$Jian Wen, Xuebo Zhang, Jing Yuan, and Yongchun Fang are with the Institute of Robotics and Automatic Information System, College of Artificial Intelligence, Nankai University, Tianjin 300350, China, and also with the Tianjin Key Laboratory of Intelligent Robotics, Nankai University, Tianjin 300350, China (e-mail: zhangxuebo@nankai.edu.cn; wenjian@mail.nankai.edu.cn).}
	\thanks{$^{2}$Qingchen Bi is with the School of Marine Science and Technology, Northwestern Polytechnical University, Xi'an 710072, China.}
	\thanks{$^{3}$Zhangchao Pan is with the Faculty of Robot Science and Engineering, Northeastern University, Shenyang 110000, China.}
	\thanks{$^{4}$Yanghe Feng is with the College of Systems Engineering, National University of Defense Technology, Changsha 410073, China.}
	\thanks{$\dagger$: Corresponding author.} %
}

\begin{document}

\maketitle
\thispagestyle{empty}
\pagestyle{empty}

\begin{abstract}
	Local planning is one of the key technologies for mobile robots to achieve full autonomy and has been widely investigated. To evaluate mobile robot local planning approaches in a unified and comprehensive way, a mobile robot local planning benchmark called MRPB 1.0 is newly proposed in this paper. The benchmark facilitates both motion planning researchers who want to compare the performance of a new local planner relative to many other state-of-the-art approaches as well as end users in the mobile robotics industry who want to select a local planner that performs best on some problems of interest. We elaborately design various simulation scenarios to challenge the applicability of local planners, including large-scale, partially unknown, and dynamic complex environments. Furthermore, three types of principled evaluation metrics are carefully designed to quantitatively evaluate the performance of local planners, wherein the safety, efficiency, and smoothness of motions are comprehensively considered. We present the application of the proposed benchmark in two popular open-source local planners to show the practicality of the benchmark. In addition, some insights and guidelines about the design and selection of local planners are also provided. The benchmark website \cite{wen2020mrpb} contains all data of the designed simulation scenarios, detailed descriptions of these scenarios, and example code. 
\end{abstract}

\section{Introduction}
\label{introduction}
Motion planning is one of the most popular research topics in mobile robotics and has been widely investigated \cite{zhang2018multilevel,chi2018risk,wang2020neural,wang2020eb}. For computational efficiency reasons, the commonly adopted motion planning framework is organized in a hierarchical architecture by combining a global planner and a local planner \cite{lunenburg2016motion}. The global planner is employed to generate a global path from the current robot pose to the goal one, followed by the local planner supposed to provide safe, flexible, and efficient motions according to real-time sensor data. In such a two-level planning scheme, the global planner only provides rough motion guidance for the robot, while the local planner plays a leading role in generating actual motions. In this work, we focus on the problem of mobile robot local planning. 

Despite mobile robot local planning approaches have been widely investigated, there is still a lack of a unified benchmark for performance evaluation. When a new local planner is proposed, the authors usually make some comparisons with other approaches through customized designed simulation or experimental scenarios. However, data sets of these designed scenarios are usually unavailable, which makes it difficult for other researchers to repeat the evaluation. Additionally, the authors may choose some specific scenarios that are friendly to their proposed approach. Therefore, it is difficult to guarantee the comprehensiveness and objectivity of the evaluation. In addition to the public data sets, there is also a lack of a complete and principled evaluation system for mobile robot local planning approaches. The use of quantitative metrics is usually limited to the total travel distance or the time taken by the robot to complete the navigation task. To make an objective performance comparison, it is necessary to use a combination of different evaluation metrics that qualify different aspects of local planners. \emph{In summary, there is a lack of public data sets for evaluating mobile robot local planning approaches, as well as principled and comprehensive evaluation metrics.}

In this paper, a mobile robot local planning benchmark called MRPB 1.0 is newly proposed to evaluate mobile robot local planning approaches in a unified and comprehensive way. We aim to establish a complete and principled evaluation framework that allows for objectively comparing the performance of local planners. To fulfill the goal, various simulation scenarios are designed and three types of evaluation metrics are proposed.

\subsubsection{Data Sets}
In order to improve the comprehensiveness and repeatability of performance evaluation, various simulation scenarios are elaborately designed and open-source. We choose Gazebo \cite{koenig2004design} as the simulation platform since its high popularity among the open-source Robot Operating System (ROS) community. On this basis, we carefully design four types of simulation scenarios, namely \emph{indoor}, \emph{narrow space}, \emph{partially unknown}, and \emph{dynamic}.

\begin{enumerate}[\hspace{1em}a)]
	\item The indoor scenarios including various-scale office-like environments are designed to make an overall evaluation of local planners.
	\item The narrow space scenarios such as complex maze environments, narrow passages, U-shaped or Z-shaped corridors, and so on are designed to challenge the flexibility and smoothness of local planners.
	\item The partially unknown scenarios, i.e., only incomplete prior information is available for local planners, are designed to pose a challenge to the adaptability of local planners in partially unknown environments. 
	\item The dynamic scenarios with moving people are designed to challenge the robustness of local planners in dealing with dynamic obstacles.
\end{enumerate}
Through these four types of simulation scenarios, the performance of local planners can be comprehensively evaluated. In addition, the repeatability of performance evaluation can be benefited from these public data sets.

\subsubsection{Metrics}
In order to comprehensively evaluate the performance of local planners from different aspects, three types of evaluation metrics are proposed, namely \emph{safety}, \emph{efficiency}, and \emph{smoothness}. 

\begin{enumerate}[\hspace{1em}a)]
	\item The safety metrics including the minimum distance to the closest obstacle and the percentage of time spent by the robot in the dangerous area around obstacles are used to evaluate the security performance of local planners.
	\item The efficiency metrics consisting of the time consumption of a single local planning and total travel time are employed to evaluate the computational efficiency and motion efficiency of local planners. 
	\item The smoothness metrics are utilized to evaluate the quality of motion commands provided by local planners, wherein the path and velocity smoothness are taken into account.
\end{enumerate}
On the basis of these three types of evaluation metrics, the performance of mobile robot local planning approaches can be evaluated in a comprehensive way.

Furthermore, we present the application of the proposed benchmark in two popular open-source local planners, namely the well-known dynamic window approach (DWA) \cite{fox1997dynamic} and the timed elastic band (TEB) local planner \cite{rosmann2013efficient} to demonstrate the practicality of the benchmark. It is shown that the optimization-based local planner TEB performs better than the sampling-based local planner DWA. On this basis, some insights and guidelines about the design and selection of local planners are provided. 

The proposed benchmark is open-source and available on our website \cite{wen2020mrpb}. \emph{This is an open work. We aim to contribute a prototype of mobile robot local planning benchmarks to the robotics community. We also look forward to making this work more complete through the open-source community.}

The main contributions of this paper are summarized as follows: 
\begin{enumerate}[\hspace{1em}1)]
	\item We present a mobile robot local planning benchmark called MRPB 1.0 to evaluate mobile robot local planning approaches in a unified and comprehensive way. A complete and principled evaluation framework that allows for objectively comparing the performance of local planners is established.
	\item A variety of simulation scenarios are elaborately designed and open-source to comprehensively evaluate the performance of local planners and benefit the repeatability of performance evaluation.
	\item Three types of principled evaluation metrics are carefully designed to comprehensively evaluate the performance of local planners from different aspects.
	\item The application of the proposed benchmark is presented. And some insights and guidelines about the design and selection of local planners are also provided. 
\end{enumerate}

We begin the paper with a brief review of related work. The designed simulation scenarios and evaluation metrics are detailed in Sections \ref{data_sets} and \ref{metrics} respectively. Section \ref{application} presents the application of the proposed benchmark and Section \ref{conclusion} comes to a conclusion.

\section{Related work}
\label{relatedwork}
Public data sets and benchmarks play an important role for scientific evaluation and objective comparison of algorithms. Previous research on mobile robot navigation benchmarks mainly focused on simultaneous localization and mapping (SLAM) techniques \cite{amigoni2018improving}, and in particular on evaluating the precision of pose estimation \cite{burgard2009comparison,kummerle2009measuring,endres2012evaluation}. In \cite{sprunk2016experimental}, Sprunk \emph{et al.} design an experimental protocol to evaluate the whole navigation system in real environments. The concept of a reference robot is introduced for comparison between different navigation systems in different experimental scenarios. In the work \cite{moll2015benchmarking}, an extensive infrastructure for analysis and visualization of sampling-based path planning algorithms is proposed and integrated into the well-known Open Motion Planning Library (OMPL) \cite{sucan2012open}. Compared with several successful benchmarks in the area of computer vision \cite{geiger2012we,sturm2012benchmark,baker2011database}, there is relatively little research on motion planning benchmarks in robotics.

In this paper, we newly propose a mobile robot local planning benchmark called MRPB 1.0 to evaluate mobile robot local planning approaches in a unified and comprehensive way. A variety of simulation scenarios are elaborately designed, taking into account large-scale, partially unknown, and dynamic complex environments. Furthermore, three types of principled evaluation metrics are proposed to comprehensively evaluate the performance of local planners from different aspects. We present the application of the proposed benchmark in two popular open-source local planners to show the practicality of the benchmark. All data of the benchmark is available on our website \cite{wen2020mrpb}.

\section{Data sets}
\label{data_sets}
To comprehensively evaluate the performance of local planners, we carefully design four types of simulation scenarios. In this section, these simulation scenarios are detailed.

\begin{figure*}[t]
	\vspace{0.2cm}
	\centering
	\subfigure[]{\includegraphics[scale=0.16]{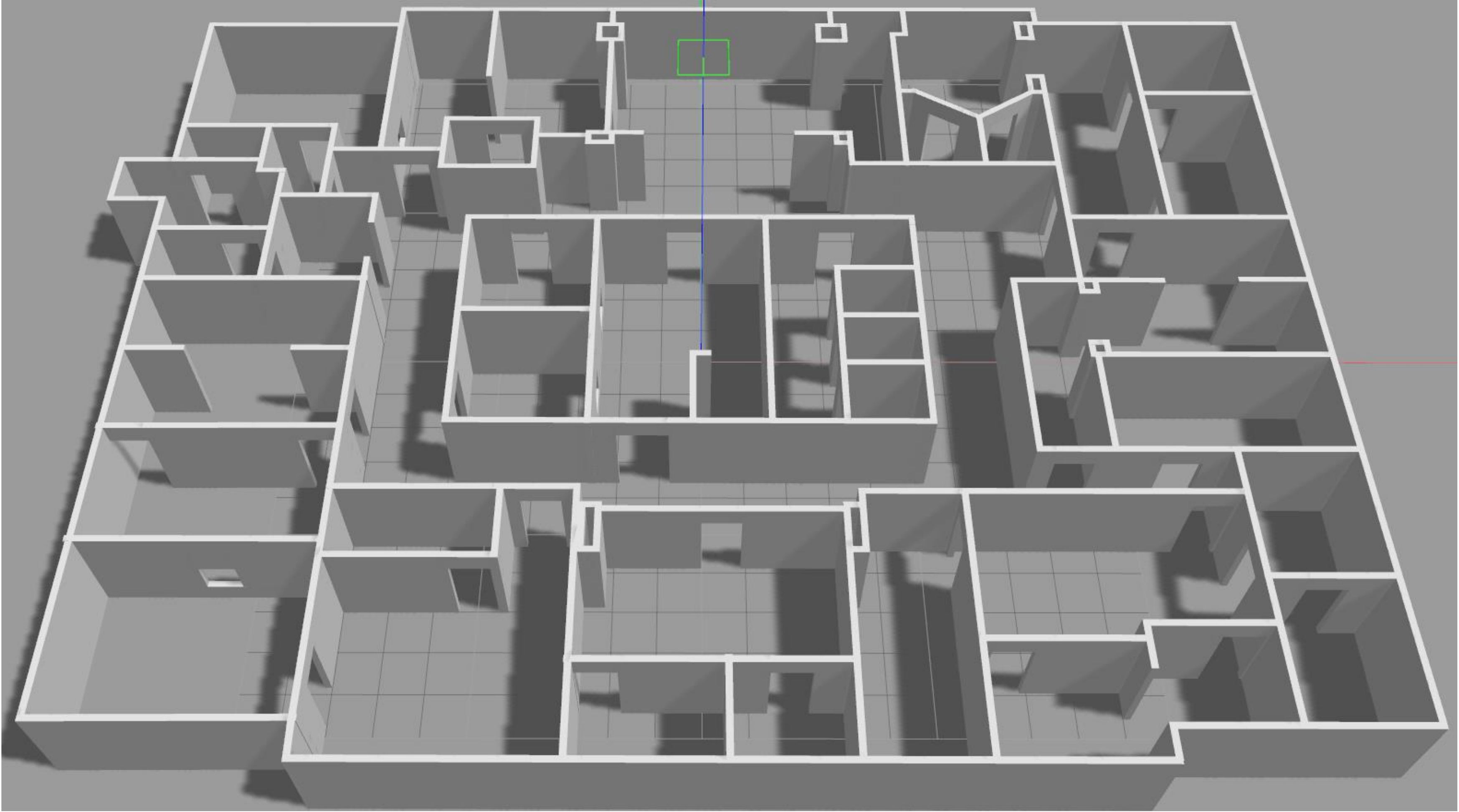}}
	\hspace{1mm}
	\subfigure[]{\includegraphics[scale=0.16]{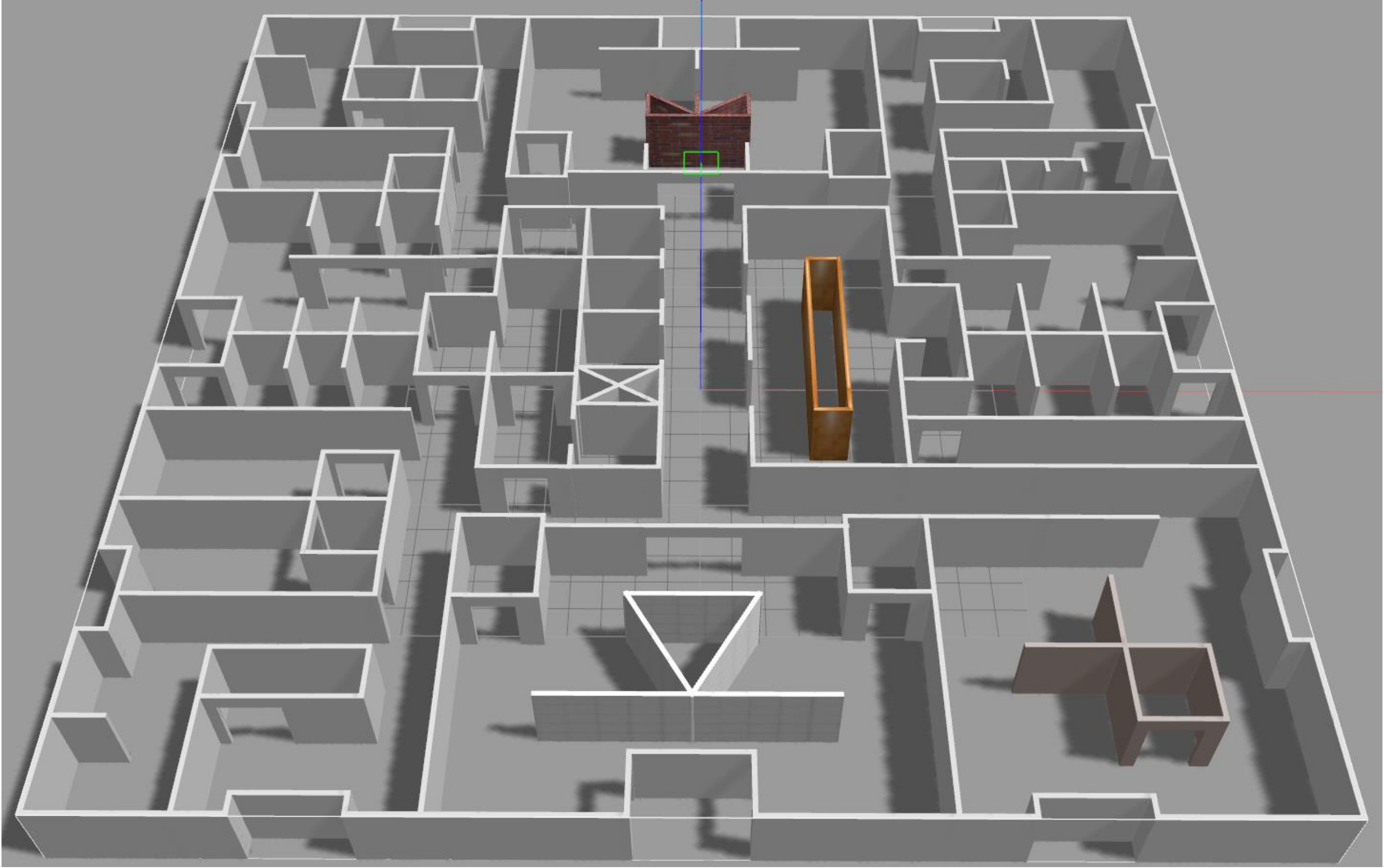}}
	\hspace{1mm}
	\subfigure[]{\includegraphics[scale=0.16]{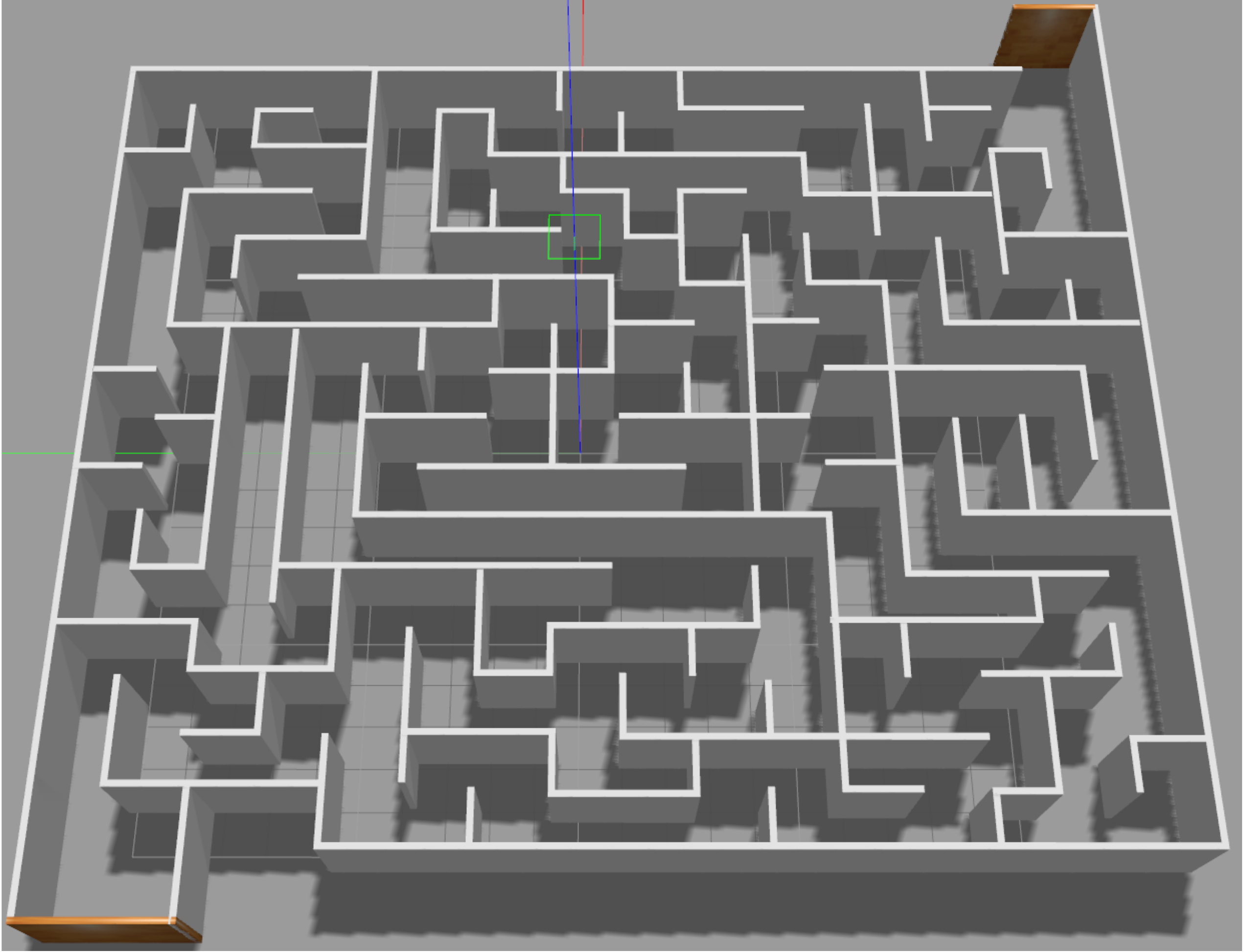}}
	\centering
	\subfigure[]{\includegraphics[scale=0.16]{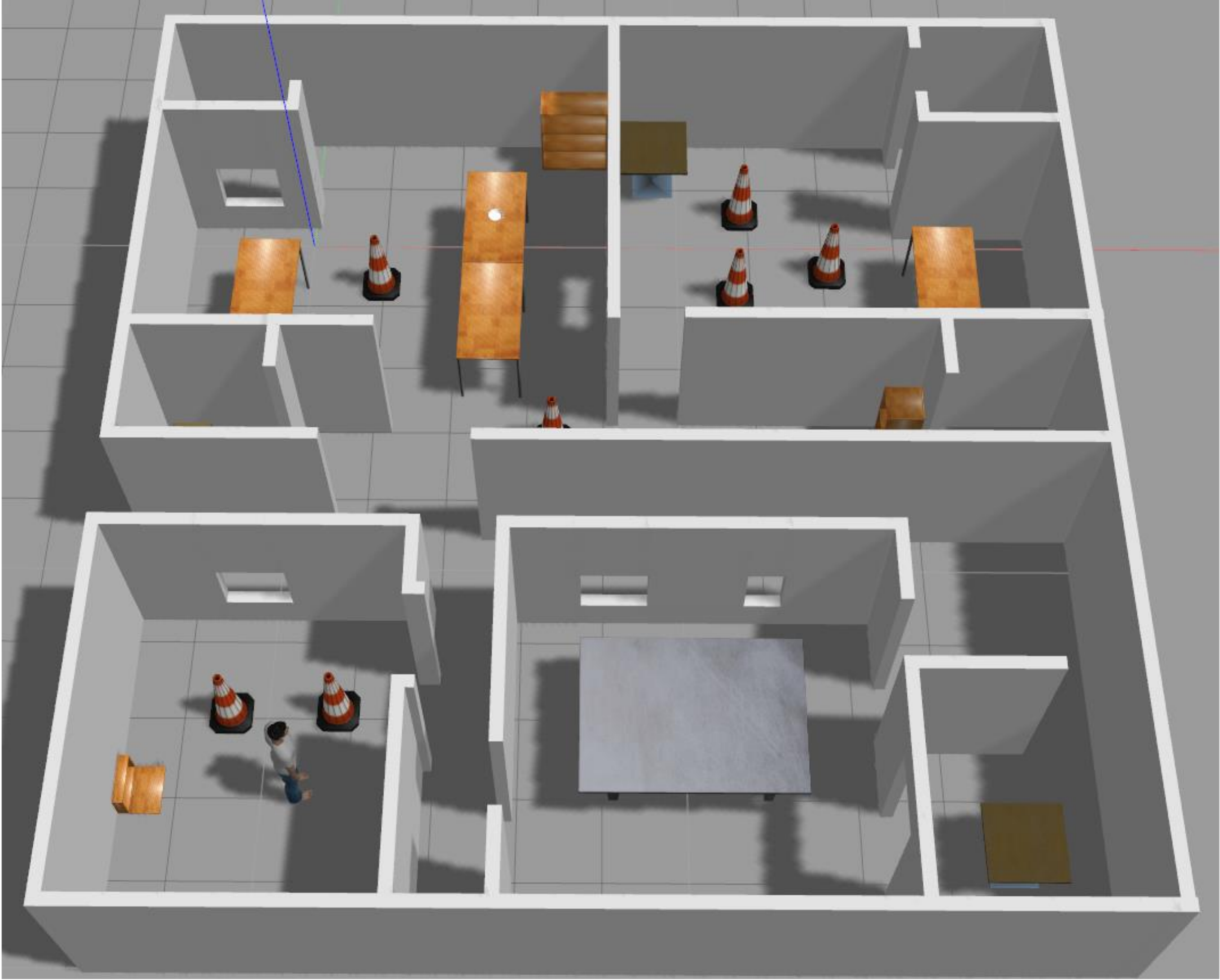}}
	\hspace{0.5mm}
	\subfigure[]{\includegraphics[scale=0.16]{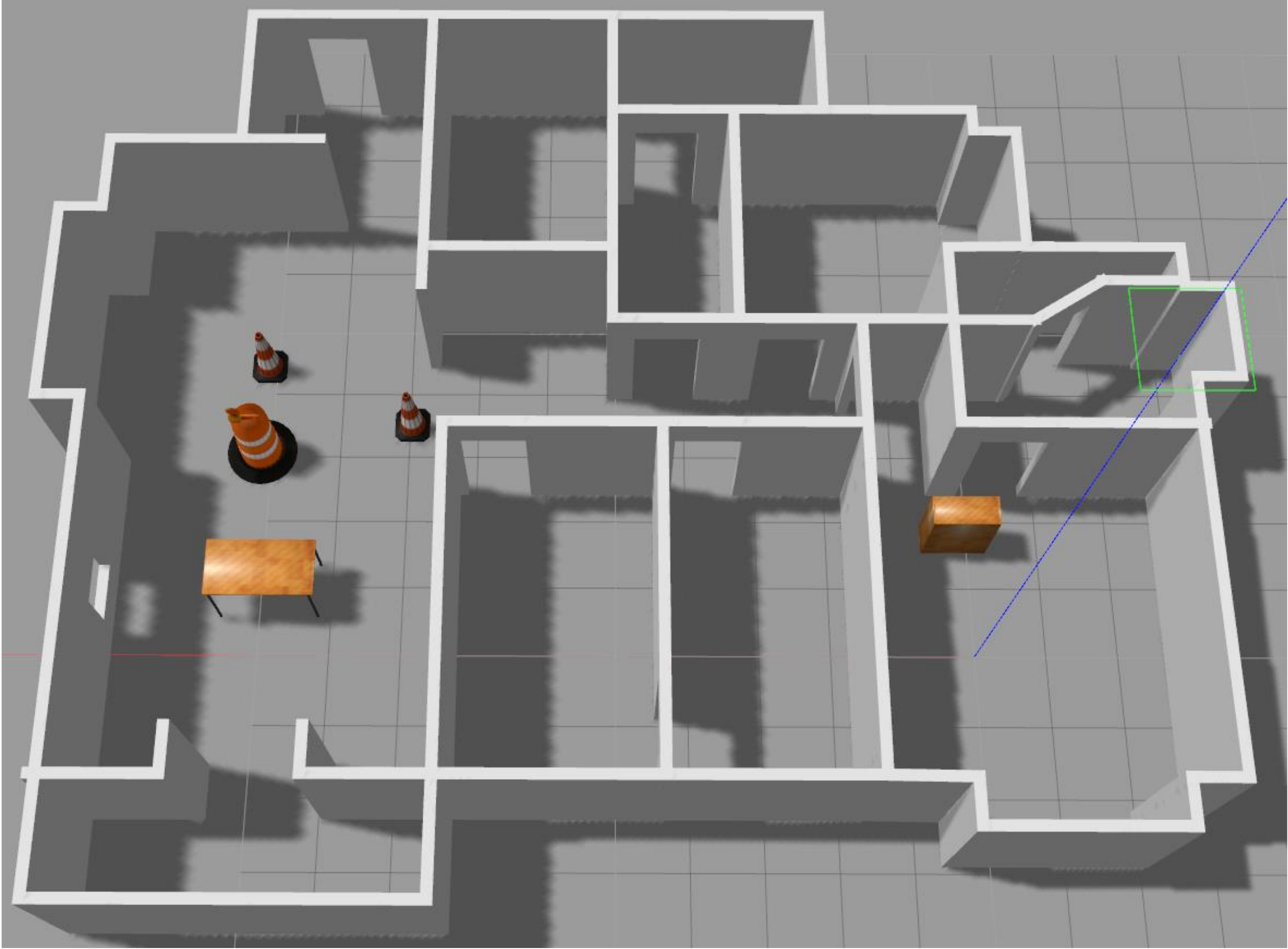}}
	\hspace{0.5mm}
	\subfigure[]{\includegraphics[scale=0.16]{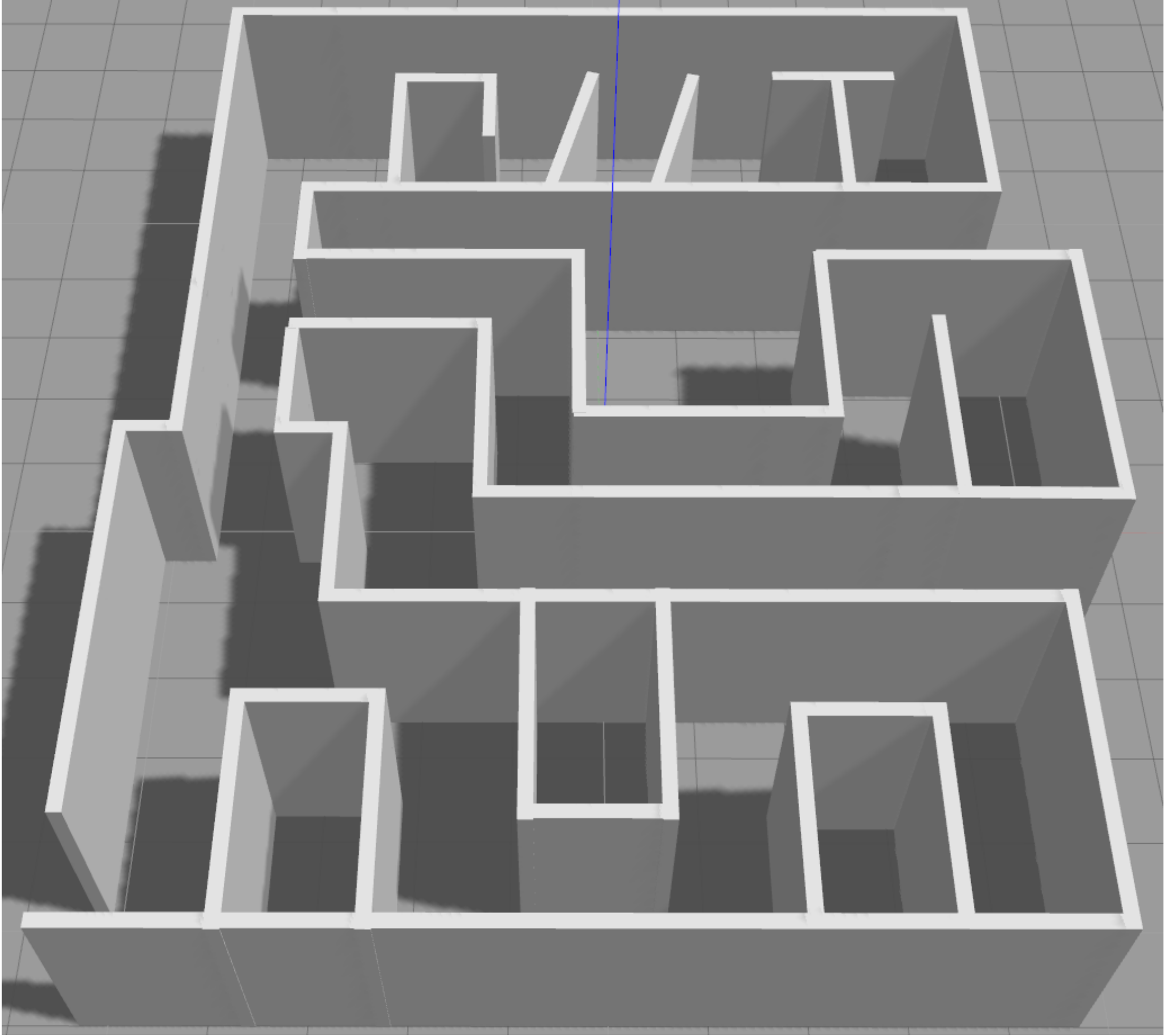}}
	\hspace{0.5mm}
	\subfigure[]{\includegraphics[scale=0.16]{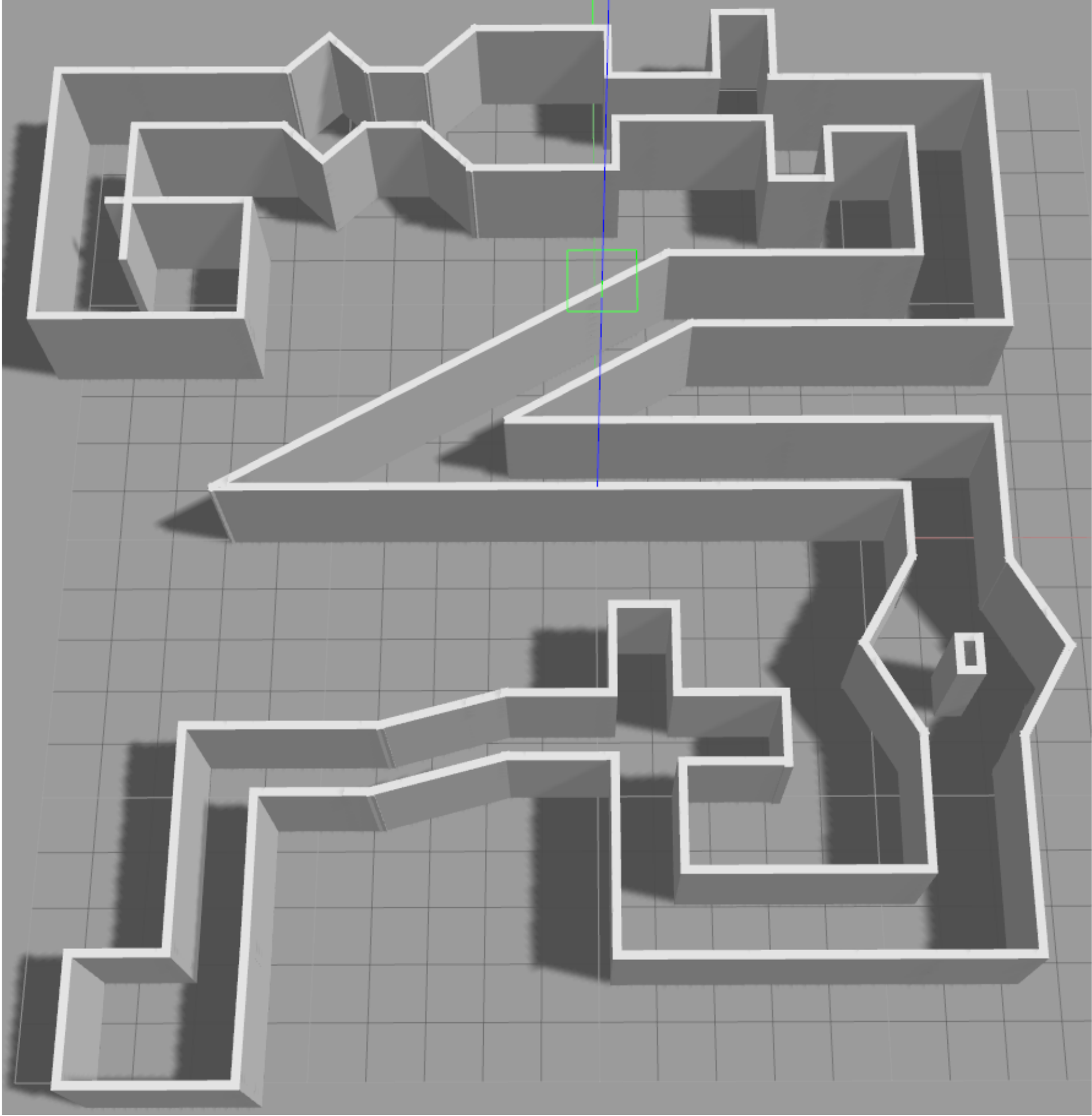}}
	\caption{Screenshots of the designed simulation scenarios. (a) Indoor office environment. (b) Shopping mall. (c) Maze. (d)-(e) Family house-like environments with dense obstacles. (f) Narrow passage with continuous U-shaped turn. (g) Narrow passage with acute-angle turn.}
	\label{datasets}
\end{figure*}

\subsection{Indoor Scenarios}
The indoor scenarios of various-scale office-like environments are designed to make an overall evaluation of local planners. The scenario shown in Fig. \ref{datasets}(a) is a $ 29.4 \times 21.9{m^2} $ indoor office environment, and the scenario depicted in Fig. \ref{datasets}(b) is designed according to the floor plan of a $ 35.2 \times 35.3{m^2} $ shopping mall. These two large-scale office-like environments are used to evaluate the applicability of local planners. In addition, two relatively small-scale family house-like scenarios are designed, as illustrated in Fig. \ref{datasets}(d) and (e). These two scenarios are used to challenge the safety and flexibility of local planners in the environment with dense obstacles.

\subsection{Narrow Space Scenarios}
In order to pose more challenges on the performance of local planners, we carefully design several narrow space scenarios. The scenario shown in Fig. \ref{datasets}(c) is a $ 23.7 \times 25.5{m^2} $ maze environment. This is an extremely challenging scenario. Firstly, the robot needs to turn continuously in the maze, requiring local planners to provide flexible motions. Secondly, the passage of the maze is relatively narrow, which requires safe motion commands to prevent the robot from colliding with the wall. In summary, the maze scenario poses a huge challenge to the flexibility and safety of local planners. 

In addition to the large-scale maze scenario, we also design two relatively small-scale narrow space scenarios. As shown in Fig. \ref{datasets}(f), a scenario with continuous U-shaped turn is designed. The robot needs to turn continuously in the narrow passage. In Fig. \ref{datasets}(g), we design an acute-angle turning scenario. At the corner, the orientation of the robot needs to be changed by approximately $ {135^ \circ } $. These scenarios are both designed to challenge the flexibility of local planners.

\begin{figure}[t]
	\centering
	\subfigure[]{\includegraphics[scale=0.25]{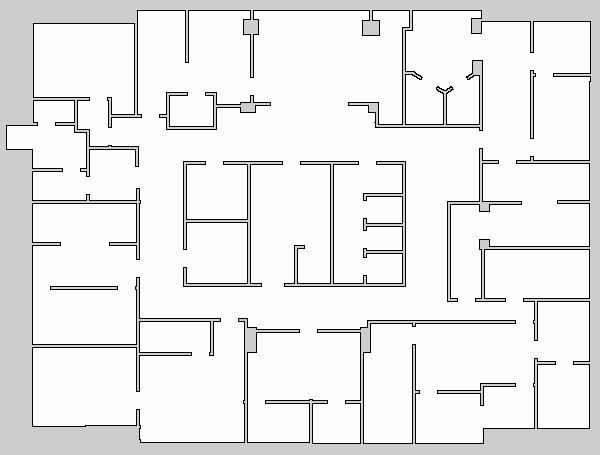}}
	\centering
	\subfigure[]{\includegraphics[scale=0.25]{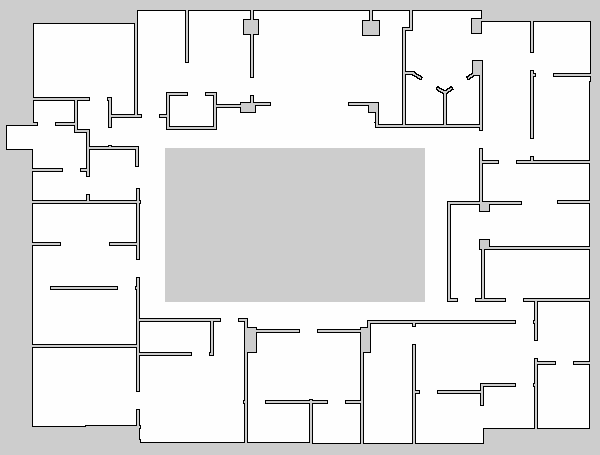}}
	\caption{Occupancy grid map of the indoor office environment. (a) Complete grid map. (b) Partially unknown grid map.}
	\label{partially_unknown}
\end{figure}

\subsection{Partially Unknown Scenarios}
In the previously designed scenarios, the complete prior map is input to local planners. To challenge the adaptability of local planners in partially unknown environments, we blur the map manually and input the incomplete map to local planners. As illustrated in Fig. \ref{partially_unknown}(a), the occupancy grid map of the indoor office environment shown in Fig. \ref{datasets}(a) is obtained by performing a wavefront exploration along the occupancy grid starting from the origin of the Gazebo world coordinate system. On this basis, we mark a rectangle of $ 13.0 \times 7.7{m^2} $ in the center of the map and set the covered cells to the unknown state, as shown in Fig. \ref{partially_unknown}(b). Then the local planning is performed in the partially unknown grid map. Local planners need to update the occupancy grid map according to real-time laser scan data and provide safe and efficient motion commands for the robot. In addition to the indoor office environment, we also design a similar scenario for the shopping mall environment.

\subsection{Dynamic Scenarios}
\begin{figure}[t]
	\vspace{0.2cm}
	\centering
	\subfigure[]{\includegraphics[scale=0.1]{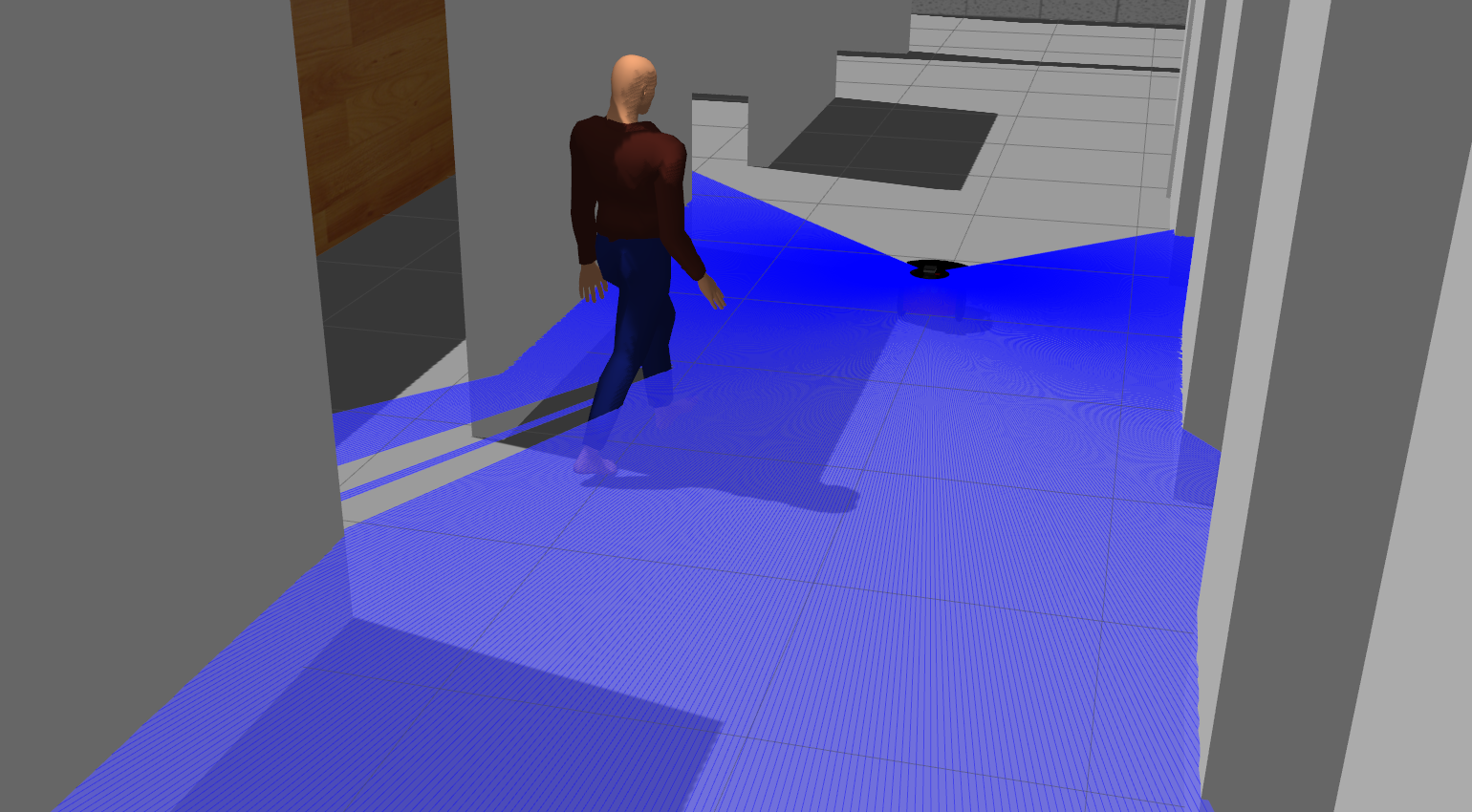}}
	\centering
	\subfigure[]{\includegraphics[scale=0.1]{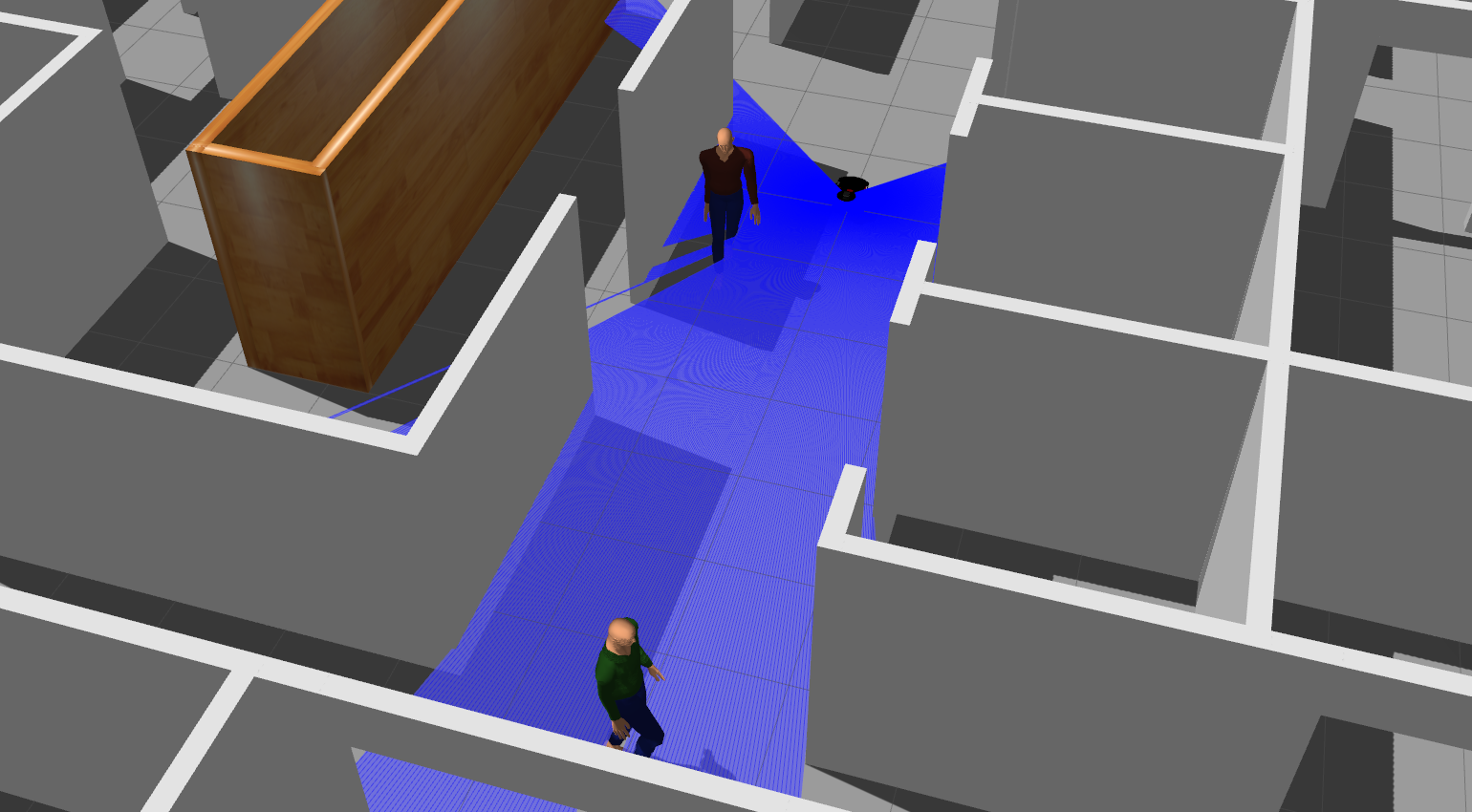}}
	\centering
	\subfigure[]{\includegraphics[scale=0.1]{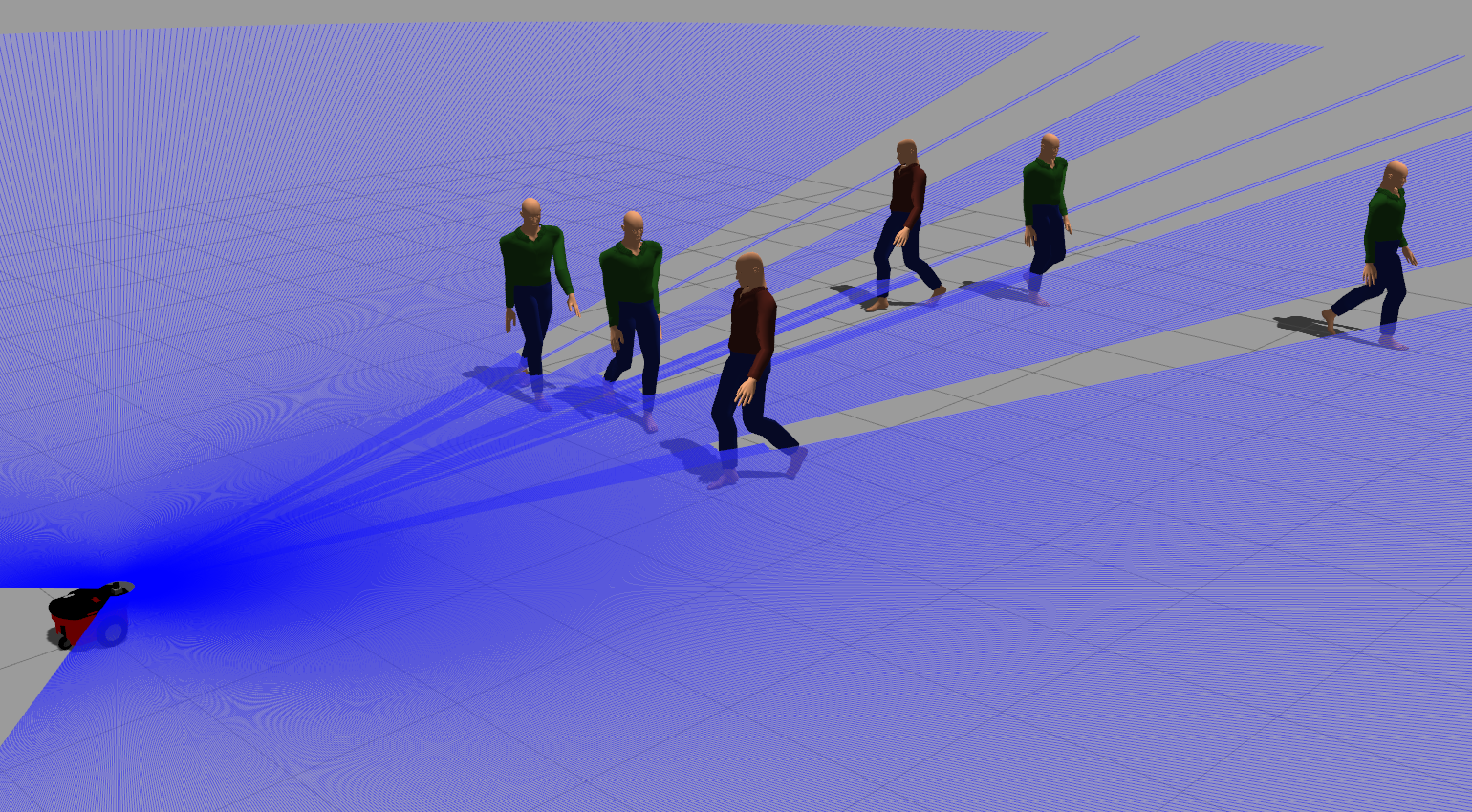}}
	\centering
	\subfigure[]{\includegraphics[scale=0.1]{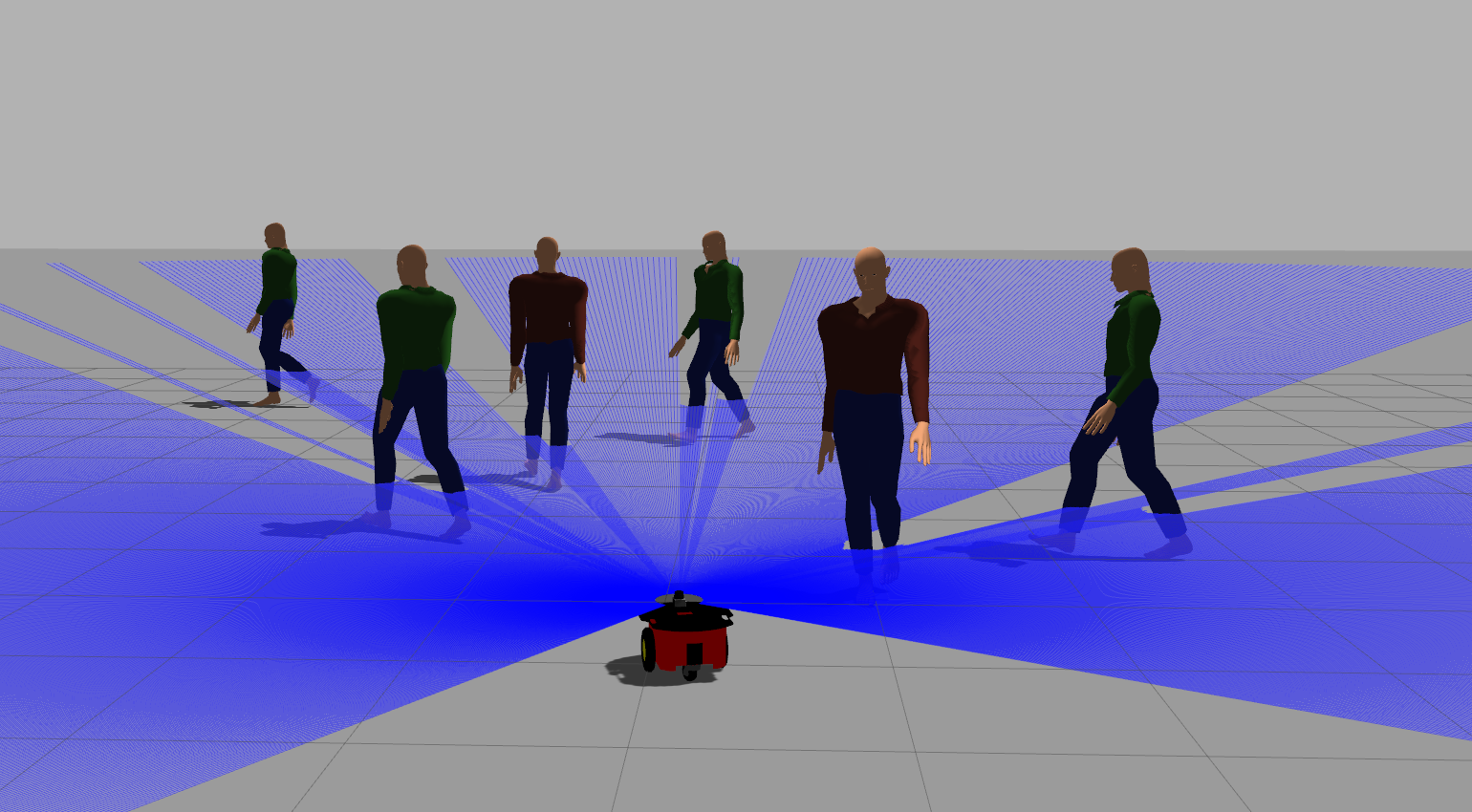}}
	\caption{Screenshots of the designed dynamic scenarios.}
	\label{dynamic}
\end{figure}

We design several dynamic scenarios to challenge the robustness of local planners in dealing with dynamic obstacles. As depicted in Fig. \ref{dynamic}(a)-(b), we simulate two people in the shopping mall environment. These two people are walking around in the T-shaped corridor at a constant speed. The robot is required to implement fast re-planning in the changing environment to avoid collisions with dynamic obstacles. We design a similar scenario for the indoor office environment. Furthermore, a more complex dynamic scenario with several moving people is designed to challenge local planners. As shown in Fig. \ref{dynamic}(c)-(d), six people are walking around in an open space environment. The robot needs to pass through the crowd to reach the goal at the other end. This scenario highly reproduces the crowded scene in the real world and poses a great challenge to the safety, flexibility, and real-time performance of local planners.

\section{Metrics}
\label{metrics}

\subsection{Data Log}
Suppose that in the process of robot navigation from the start pose to the goal one, the local planning is called $ N $ times. Every time the local planning is called, we need to log the following data: the timestamp of the $ i $-th call $ {{t_i}} $, the robot pose $ \left( {{x_i},{y_i},{\theta _i}} \right) $, the linear and angular velocities $ \left( {{v_i},{\omega _i}} \right) $, the distance to the closest obstacle $ {d_i} $, and the time consumption of the local planning $ {c_i} $. When the robot completes the navigation task, we obtain the intermediate data of the whole navigation process $ \left\{ {{t_i},{x_i},{y_i},{\theta _i},{v_i},{\omega _i},{d_i},{c_i}} \right\},1 \le i \le N $.

\subsection{Safety Metrics}
The safety metrics are employed to evaluate the security performance of local planners in guiding the robot to the goal. In this work, the minimum distance to the closest obstacle $ {d_o} $ and the percentage of time spent by the robot in the dangerous area around obstacles $ {p_o} $ are used to evaluate the security of local planners
\begin{equation}
	{d_o} = \min \left\{ {{d_i}} \right\},1 \le i \le N,
\end{equation}
\begin{equation}
	{p_o} = \frac{{\sum {\left( {{t_b} - {t_a}} \right)} }}{{{t_N} - {t_1}}} \times 100\%,
\end{equation}
where the subscripts $ a $ and $ b $ are the indices of timestamps satisfying $ {d_k} \le {d_{{\rm{safe}}}},a \le k \le b $, and $ {d_{{\rm{safe}}}} $ is the preset safe distance to obstacles. The distance to the closest obstacle $ {d_i} $ is obtained by combining an efficient distance transform algorithm described in \cite{felzenszwalb2012distance} and bicubic interpolation on top of the occupancy grid map.

\subsection{Efficiency Metrics}
The efficiency metrics are used to evaluate the motion efficiency and computational efficiency of local planners. The motion efficiency measures how quickly the local planner guides the robot to the goal, and the computational efficiency evaluates the real-time performance of local planners. In this work, the total travel time $ T $ of the robot from the start to the goal is used to evaluate the motion efficiency
\begin{equation}
	T = {t_N} - {t_1}.
\end{equation}
And the computational efficiency is measured as the average period between the time when the local planner receives the planning request and the time of the updated twist command becoming available
\begin{equation}
	C = \frac{1}{N}\sum\limits_{i = 1}^N {{c_i}}.
\end{equation}

\begin{table}[t]
	\vspace{0.2cm}
	\centering
	\caption{Common parameters of local planners for the evaluation}
	\label{parameter}
	\scalebox{0.9}{
	\begin{tabular}{cccccccccc}
		\toprule
		Parameter	& $ {\bar v_{\max }}/{\bar v_{\min }} $		& $ {\bar \omega _{\max }}/{\bar \omega _{\min }} $		& $ {\bar a_{\max }}/{\bar a_{\min }} $ 	& $ {\bar \alpha _{\max }}/{\bar \alpha _{\min }} $	 \\
		\midrule
		Value		& $ 0.55/ - 0.2 $		& $ 1.0 / -1.0 $	& $ 2.5 / -2.5 $	& $ 3.2 / -3.2 $			\\
		Unit		& $ m/s $ 				& $ rad/s $ 		& $ m/{s^2} $ 		& $ rad/{s^2} $				\\
		\bottomrule
	\end{tabular}}
\end{table}

\subsection{Smoothness Metrics}
The smoothness metrics are employed to evaluate the quality of motion commands provided by local planners. In this work, the smoothness performance of local planners is comprehensively evaluated by the path smoothness and velocity smoothness. We broadly follow the smoothness constraint defined in \cite{dolgov2009autonomous} to evaluate the path smoothness
\begin{equation}
	{f_{ps}} = \sum\limits_{i = 2}^{N - 1} {{{\left\| {{\bf{\Delta}} {{\bf{x}}_{i + 1}} - {\bf{\Delta}} {{\bf{x}}_i}} \right\|}^{\rm{2}}}},
\end{equation}
where $ {\bf{\Delta}} {{\bf{x}}_i} = {{\bf{x}}_i} - {{\bf{x}}_{i - 1}},2 \le i \le N $ denotes the displacement vector at the vertex $ {{\bf{x}}_i} = {\left( {{x_i},{y_i}} \right)^{\rm{T}}} $. And the velocity smoothness is measured by the average of the acceleration
\begin{equation}
	{f_{vs}} = \frac{1}{{N - 1}}\sum\limits_{i = 1}^{N - 1} {\left| {\frac{{{v_{i + 1}} - {v_i}}}{{{t_{i + 1}} - {t_i}}}} \right|}.
\end{equation}

\section{Application}
\label{application}

\subsection{Setup}
As mentioned before, we choose Gazebo as the simulation platform since its high popularity among the ROS community. Currently, the benchmark is released with ROS Kinetic and Gazebo 9. All the evaluations are performed on a laptop with an Intel Core i5-7200U processor and 8 GB RAM.

\begin{table*}[t]
	\vspace{0.2cm}
	\renewcommand\arraystretch{1.3}
	\centering
	\caption{Quantitative statistics of local planning results in static environments}
	\label{statictable}
	\scalebox{0.9}{
	\begin{tabular}{|c|c|c|c|c|c|c|c|c|c|c|c|c|c|c|c|}
		\hline
		\multicolumn{2}{|c|}{ \multirow{2}*{} } & \multicolumn{4}{c|}{\textbf{Safety}} &\multicolumn{4}{c|}{\textbf{Efficiency}} &\multicolumn{4}{c|}{\textbf{Smoothness}} & \multicolumn{2}{c|}{\textbf{Path length}}  \\
		\cline{3-16}
		\multicolumn{2}{|c|}{} & \multicolumn{2}{c|}{$ {d_o}\left[ m \right] $} & \multicolumn{2}{c|}{$ {p_o}\left[ \%  \right] $} & \multicolumn{2}{c|}{$ T\left[ s \right] $}  & \multicolumn{2}{c|}{$ C\left[ ms \right] $} & \multicolumn{2}{c|}{$ {f_{ps}}\left[ {{m^2}} \right] $} & \multicolumn{2}{c|}{$ {f_{vs}}\left[ {m/{s^2}} \right] $}  & \multicolumn{2}{c|}{$ S\left[ m \right] $} \\
		\cline{3-16}
		\multicolumn{2}{|c|}{} & \textsl{\textbf{DWA}} & \textsl{\textbf{TEB}} & \textsl{\textbf{DWA}} & \textsl{\textbf{TEB}} & \textsl{\textbf{DWA}} & \textsl{\textbf{TEB}} & \textsl{\textbf{DWA}} & \textsl{\textbf{TEB}} & \textsl{\textbf{DWA}} & \textsl{\textbf{TEB}} & \textsl{\textbf{DWA}} & \textsl{\textbf{TEB}} & \textsl{\textbf{DWA}} & \textsl{\textbf{TEB}} \\
		\cline{1-16}
		\multicolumn{1}{|c|}{ \multirow{5}*{\textbf{Scenario (a)}} } & 1 & 0.20 & \textbf{0.29} & 10.1 & \textbf{1.1} & 38.5 & \textbf{35.0} & 46.7 & \textbf{3.7} & 0.05 & \textbf{0.02} & 0.16 & \textbf{0.03} & \textbf{18.65} & 18.84 \\
		\cline{2-16}
		\multicolumn{1}{|c|}{} & 2 & 0.20 & \textbf{0.29} & 16.0 & \textbf{4.8} & 47.7 & \textbf{42.0} & 41.8 & \textbf{5.2} & 0.06 & \textbf{0.03} & 0.15 & \textbf{0.04} & \textbf{21.48} & 22.03 \\
		\cline{2-16}
		\multicolumn{1}{|c|}{} & 3 & 0.20 & \textbf{0.28} & 7.9 & \textbf{1.6} & \textbf{74.1} & 76.8 & 41.4 & \textbf{4.1} & 0.10 & \textbf{0.03} & 0.10 & \textbf{0.03} & \textbf{36.66} & 37.20 \\
		\cline{2-16}
		\multicolumn{1}{|c|}{} & 4 & 0.20 & \textbf{0.27} & 10.8 & \textbf{3.8} & 56.6 & \textbf{52.6} & 42.1 & \textbf{3.8} & 0.52 & \textbf{0.02} & 0.09 & \textbf{0.03} & \textbf{28.49} & 28.69 \\
		\cline{2-16}
		\multicolumn{1}{|c|}{} & 5 & 0.20 & \textbf{0.28} & 10.3 & \textbf{3.4} & 76.4 & \textbf{58.4} & 37.7 & \textbf{4.8} & 0.66 & \textbf{0.03} & 0.09 & \textbf{0.03} & 31.71 & \textbf{30.87} \\
		\cline{1-16}
		\multicolumn{1}{|c|}{ \multirow{5}*{\textbf{Scenario (b)}} } & 1 & 0.22 & \textbf{0.28} & 4.5 & \textbf{2.9} & 84.2 & \textbf{81.4} & 43.0 & \textbf{4.2} & 0.32 & \textbf{0.03} & 0.05 & \textbf{0.03} & \textbf{43.24} & 43.45 \\
		\cline{2-16}
		\multicolumn{1}{|c|}{} & 2 & 0.20 & \textbf{0.27} & 9.4 & \textbf{2.1} & 127.2 & \textbf{103.6} & 39.1 & \textbf{4.4} & 0.16 & \textbf{0.04} & 0.07 & \textbf{0.02} & 57.62 & \textbf{56.99} \\
		\cline{2-16}
		\multicolumn{1}{|c|}{} & 3 & \textbf{0.40} & 0.32 & \textbf{0.0} & 0.7 & 69.2 & \textbf{59.8} & 39.9 & \textbf{4.1} & 3.84 & \textbf{0.02} & 0.04 & \textbf{0.02} & 32.55 & \textbf{32.25} \\
		\cline{2-16}
		\multicolumn{1}{|c|}{} & 4 & 0.28 & \textbf{0.29} & \textbf{0.4} & 0.8 & 83.0 & \textbf{77.2} & 40.9 & \textbf{4.1} & 7.12 & \textbf{0.04} & 0.04 & \textbf{0.03} & 42.28 & \textbf{42.17} \\
		\cline{2-16}
		\multicolumn{1}{|c|}{} & 5 & 0.28 & \textbf{0.29} & \textbf{0.6} & 2.6 & 63.6 & \textbf{61.6} & 44.2 & \textbf{4.1} & 2.79 & \textbf{0.03} & 0.04 & \textbf{0.01} & 34.27 & \textbf{33.96} \\
		\cline{1-16}
		\multicolumn{1}{|c|}{ \multirow{5}*{\textbf{Scenario (c)}} } & 1 & 0.20 & \textbf{0.29} & 20.3 & \textbf{3.5} & 102.7 & \textbf{78.8} & 41.2 & \textbf{5.4} & 0.42 & \textbf{0.05} & 0.17 & \textbf{0.03} & 42.02 & 42.02 \\
		\cline{2-16}
		\multicolumn{1}{|c|}{} & 2 & 0.20 & \textbf{0.27} & 33.3 & \textbf{4.8} & 82.0 & \textbf{58.6} & 39.8 & \textbf{4.9} & 0.26 & \textbf{0.04} & 0.15 & \textbf{0.05} & 32.72 & \textbf{31.86} \\
		\cline{2-16}
		\multicolumn{1}{|c|}{} & 3 & 0.20 & \textbf{0.29} & 21.0 & \textbf{4.4} & 77.1 & \textbf{67.8} & 43.8 & \textbf{5.4} & 0.34 & \textbf{0.05} & 0.17 & \textbf{0.02} & 36.93 & \textbf{36.71} \\
		\cline{2-16}
		\multicolumn{1}{|c|}{} & 4 & - & \textbf{0.28} & - & \textbf{4.2} & - & \textbf{61.8} & - & \textbf{4.1} & - & \textbf{0.04} & - & \textbf{0.03} & - & \textbf{27.68} \\
		\cline{2-16}
		\multicolumn{1}{|c|}{} & 5 & - & \textbf{0.25} & - & \textbf{9.5} & - & \textbf{40.0} & - & \textbf{6.3} & - & \textbf{0.08} & - & \textbf{0.08} & - & \textbf{20.09} \\
		\cline{1-16}
		\multicolumn{1}{|c|}{ \multirow{3}*{\textbf{Scenario (d)}} } & 1 & 0.20 & \textbf{0.29} & \textbf{50.9} & 96.4 & 90.2 & \textbf{33.4} & 20.3 & \textbf{4.6} & 0.10 & \textbf{0.02} & 0.12 & \textbf{0.04} & 16.53 & \textbf{16.12} \\
		\cline{2-16}
		\multicolumn{1}{|c|}{} & 2 & 0.20 & \textbf{0.30} & 27.3 & \textbf{7.9} & 31.7 & \textbf{25.4} & 35.0 & \textbf{6.2} & 0.12 & \textbf{0.02} & 0.18 & \textbf{0.05} & 13.30 & \textbf{12.86} \\
		\cline{2-16}
		\multicolumn{1}{|c|}{} & 3 & 0.20 & \textbf{0.29} & 42.7 & \textbf{7.7} & 39.1 & \textbf{31.2} & 25.6 & \textbf{3.7} & 0.13 & \textbf{0.03} & 0.26 & \textbf{0.05} & \textbf{11.99} & 12.19 \\
		\cline{1-16}
		\multicolumn{1}{|c|}{ \multirow{3}*{\textbf{Scenario (e)}} } & 1 & 0.20 & \textbf{0.29} & 16.4 & \textbf{6.1} & 31.8 & \textbf{26.2} & 42.3 & \textbf{4.4} & 0.03 & \textbf{0.02} & 0.17 & \textbf{0.04} & 14.47 & \textbf{14.34} \\
		\cline{2-16}
		\multicolumn{1}{|c|}{} & 2 & 0.20 & \textbf{0.32} & 3.8 & \textbf{0.8} & 27.9 & \textbf{25.4} & 49.6 & \textbf{3.7} & 0.05 & \textbf{0.02} & 0.10 & \textbf{0.03} & \textbf{13.79} & 13.89 \\
		\cline{2-16}
		\multicolumn{1}{|c|}{} & 3 & - & \textbf{0.29} & - & \textbf{3.3} & - & \textbf{24.6} & - & \textbf{5.1} & - & \textbf{0.02} & - & \textbf{0.07} & - & \textbf{11.80} \\
		\cline{1-16}
		\multicolumn{1}{|c|}{ \multirow{3}*{\textbf{Scenario (f)}} } & 1 & 0.20 & \textbf{0.27} & 19.9 & \textbf{4.2} & 53.0 & \textbf{47.4} & 44.0 & \textbf{6.1} & \textbf{0.05} & 0.06 & 0.15 & \textbf{0.07} & \textbf{24.18} & 24.53 \\
		\cline{2-16}
		\multicolumn{1}{|c|}{} & 2 & \textbf{0.20} & 0.18 & 38.9 & \textbf{12.4} & 51.4 & \textbf{46.8} & 41.9 & \textbf{6.3} & \textbf{0.12} & 0.14 & 0.18 & \textbf{0.10} & \textbf{22.63} & 22.98 \\
		\cline{2-16}
		\multicolumn{1}{|c|}{} & 3 & - & \textbf{0.28} & - & \textbf{8.1} & - & \textbf{27.0} & - & \textbf{4.9} & - & \textbf{0.03} & - & \textbf{0.06} & - & \textbf{14.00} \\
		\cline{1-16}
		\multicolumn{1}{|c|}{\textbf{Scenario (g)}} & 1 & \textbf{0.22} & 0.20 & 7.9 & \textbf{4.9} & \textbf{126.4} & 129.4 & 53.4 & \textbf{5.7} & 0.86 & \textbf{0.17} & \textbf{0.03} & 0.05 & 68.33 & \textbf{68.09} \\
		\hline
	\end{tabular}}
\end{table*}

In the evaluation, we focus on the local planning problem of differential drive mobile robots. We refer to the popular Pioneer 3-DX mobile robot and design a robot model in Gazebo through the XML macros language Xacro. The footprint of the robot is broadly set to a circle with a radius of $ 0.17m $. On this basis, a laser rangefinder is mounted on the robot. The laser rangefinder has a scanning range of $ - {135^ \circ } \sim {135^ \circ }$ with the angular resolution being $0.25^\circ$, and the effective measurement range is $0.1m \sim 30m$.

The navigation tests are performed based on the powerful ROS navigation stack \cite{marder2010office}. To make a fair comparison between different local planners, the global planner proposed in \cite{konolige2000gradient} is used in all tests, and some common parameters such as the maximum and minimum linear velocities are also set to the same. The settings of these parameters are enumerated in Table \ref{parameter}, wherein $ \bar v $, $ \bar \omega $, $ \bar a $, and $ \bar \alpha $ denote the linear velocity, angular velocity, linear acceleration, and angular acceleration respectively. The period of the local planning thread is set to $ 0.2s $, and the safe distance to obstacles is set to $ 0.34m $, i.e., twice the radius of the robot. Considering the computational efficiency, the local planning is performed in a $ 5.5 \times 5.5{m^2} $ local map with the resolution being $ 0.1m/cell $. In addition, the robot pose is obtained from the ground truth provided by Gazebo to avoid the influence of localization error. To roughly describe the distance traveled by the robot during the navigation process, we also present the path length in the evaluation results. The two local planners of DWA\footnote{{http://wiki.ros.org/dwa\_local\_planner}} and TEB\footnote{{http://wiki.ros.org/teb\_local\_planner}} each have some specific parameters to be set. Except for the common parameters enumerated in Table \ref{parameter}, we use their own default parameters for the evaluation. Readers are advised to refer to the original papers \cite{fox1997dynamic,rosmann2013efficient} for more details about these two local planners.


\begin{table*}[t]
	\vspace{0.2cm}
	\renewcommand\arraystretch{1.3}
	\centering
	\caption{Quantitative statistics of local planning results in partially unknown environments}
	\label{unknowntable}
	\scalebox{0.9}{
		\begin{tabular}{|c|c|c|c|c|c|c|c|c|c|c|c|c|c|c|c|}
			\hline
			\multicolumn{2}{|c|}{ \multirow{2}*{} } & \multicolumn{4}{c|}{\textbf{Safety}} &\multicolumn{4}{c|}{\textbf{Efficiency}} &\multicolumn{4}{c|}{\textbf{Smoothness}} & \multicolumn{2}{c|}{\textbf{Path length}}  \\
			\cline{3-16}
			\multicolumn{2}{|c|}{} & \multicolumn{2}{c|}{$ {d_o}\left[ m \right] $} & \multicolumn{2}{c|}{$ {p_o}\left[ \%  \right] $} & \multicolumn{2}{c|}{$ T\left[ s \right] $}  & \multicolumn{2}{c|}{$ C\left[ ms \right] $} & \multicolumn{2}{c|}{$ {f_{ps}}\left[ {{m^2}} \right] $} & \multicolumn{2}{c|}{$ {f_{vs}}\left[ {m/{s^2}} \right] $}  & \multicolumn{2}{c|}{$ S\left[ m \right] $} \\
			\cline{3-16}
			\multicolumn{2}{|c|}{} & \textsl{\textbf{DWA}} & \textsl{\textbf{TEB}} & \textsl{\textbf{DWA}} & \textsl{\textbf{TEB}} & \textsl{\textbf{DWA}} & \textsl{\textbf{TEB}} & \textsl{\textbf{DWA}} & \textsl{\textbf{TEB}} & \textsl{\textbf{DWA}} & \textsl{\textbf{TEB}} & \textsl{\textbf{DWA}} & \textsl{\textbf{TEB}} & \textsl{\textbf{DWA}} & \textsl{\textbf{TEB}} \\
			\cline{1-16}
			\multicolumn{1}{|c|}{ \multirow{5}*{\textbf{Scenario (a)}} } & 1 & 0.22 & \textbf{0.32} & 4.9 & \textbf{1.2} & 37.5 & \textbf{34.0} & 42.9 & \textbf{3.9} & 0.25 & \textbf{0.02} & 0.06 & \textbf{0.02} & 18.58 & \textbf{18.57} \\
			\cline{2-16}
			\multicolumn{1}{|c|}{} & 2 & 0.22 & \textbf{0.28} & 5.6 & \textbf{3.5} & 61.2 & \textbf{57.0} & 42.9 & \textbf{4.2} & 1.38 & \textbf{0.02} & 0.05 & \textbf{0.02} & \textbf{30.70} & 30.87 \\
			\cline{2-16}
			\multicolumn{1}{|c|}{} & 3 & 0.20 & \textbf{0.28} & 15.5 & \textbf{2.9} & 44.5 & \textbf{41.8} & 44.7 & \textbf{3.9} & 0.54 & \textbf{0.02} & 0.10 & \textbf{0.02} & \textbf{22.86} & 22.96 \\
			\cline{2-16}
			\multicolumn{1}{|c|}{} & 4 & 0.22 & \textbf{0.30} & 10.1 & \textbf{1.3} & 53.7 & \textbf{46.6} & 41.6 & \textbf{4.1} & 0.18 & \textbf{0.04} & 0.16 & \textbf{0.05} & 25.61 & \textbf{24.25} \\
			\cline{2-16}
			\multicolumn{1}{|c|}{} & 5 & 0.20 & \textbf{0.27} & 4.3 & \textbf{2.7} & 65.5 & \textbf{59.4} & 40.5 & \textbf{4.8} & 0.92 & \textbf{0.03} & 0.09 & \textbf{0.02} & 30.99 & \textbf{30.83} \\
			\cline{1-16}
			\multicolumn{1}{|c|}{ \multirow{5}*{\textbf{Scenario (b)}} } & 1 & \textbf{0.32} & 0.28 & \textbf{0.0} & 1.4 & 49.9 & \textbf{43.4} & 39.8 & \textbf{3.2} & 3.79 & \textbf{0.02} & 0.07 & \textbf{0.03} & \textbf{23.55} & 23.58 \\
			\cline{2-16}
			\multicolumn{1}{|c|}{} & 2 & 0.28 & \textbf{0.29} & 1.5 & \textbf{0.9} & 97.2 & \textbf{67.6} & 41.7 & \textbf{4.3} & 3.10 & \textbf{0.05} & 0.10 & \textbf{0.03} & 40.32 & \textbf{36.02} \\
			\cline{2-16}
			\multicolumn{1}{|c|}{} & 3 & 0.28 & \textbf{0.32} & \textbf{0.0} & 0.7 & 75.9 & \textbf{59.8} & 39.9 & \textbf{4.0} & 4.46 & \textbf{0.02} & 0.05 & \textbf{0.02} & 32.81 & \textbf{32.27} \\
			\cline{2-16}
			\multicolumn{1}{|c|}{} & 4 & \textbf{0.32} & 0.28 & \textbf{0.1} & 2.8 & 113.0 & \textbf{105.4} & 42.0 & \textbf{4.6} & 7.34 & \textbf{0.05} & 0.04 & \textbf{0.02} & 58.18 & \textbf{57.84} \\
			\cline{2-16}
			\multicolumn{1}{|c|}{} & 5 & 0.20 & \textbf{0.25} & \textbf{2.5} & 2.6 & 65.0 & \textbf{61.6} & 39.4 & \textbf{3.8} & 0.86 & \textbf{0.04} & 0.09 & \textbf{0.05} & \textbf{31.72} & 32.34 \\
			\cline{1-16}
			\hline
	\end{tabular}}
\end{table*}

\begin{table*}[t]
	\renewcommand\arraystretch{1.3}
	\centering
	\caption{Quantitative statistics of local planning results in dynamic environments}
	\label{dynamictable}
	\scalebox{0.9}{
		\begin{tabular}{|c|c|c|c|c|c|c|c|c|c|c|c|c|c|c|c|}
			\hline
			\multicolumn{2}{|c|}{ \multirow{2}*{} } & \multicolumn{4}{c|}{\textbf{Safety}} &\multicolumn{4}{c|}{\textbf{Efficiency}} &\multicolumn{4}{c|}{\textbf{Smoothness}} & \multicolumn{2}{c|}{\textbf{Path length}}  \\
			\cline{3-16}
			\multicolumn{2}{|c|}{} & \multicolumn{2}{c|}{$ {d_o}\left[ m \right] $} & \multicolumn{2}{c|}{$ {p_o}\left[ \%  \right] $} & \multicolumn{2}{c|}{$ T\left[ s \right] $}  & \multicolumn{2}{c|}{$ C\left[ ms \right] $} & \multicolumn{2}{c|}{$ {f_{ps}}\left[ {{m^2}} \right] $} & \multicolumn{2}{c|}{$ {f_{vs}}\left[ {m/{s^2}} \right] $}  & \multicolumn{2}{c|}{$ S\left[ m \right] $} \\
			\cline{3-16}
			\multicolumn{2}{|c|}{} & \textsl{\textbf{DWA}} & \textsl{\textbf{TEB}} & \textsl{\textbf{DWA}} & \textsl{\textbf{TEB}} & \textsl{\textbf{DWA}} & \textsl{\textbf{TEB}} & \textsl{\textbf{DWA}} & \textsl{\textbf{TEB}} & \textsl{\textbf{DWA}} & \textsl{\textbf{TEB}} & \textsl{\textbf{DWA}} & \textsl{\textbf{TEB}} & \textsl{\textbf{DWA}} & \textsl{\textbf{TEB}} \\
			\cline{1-16}
			\multicolumn{1}{|c|}{ \multirow{5}*{\textbf{Scenario (a)}} } & 1 & 0.22 & \textbf{0.32} & 3.7 & \textbf{0.7} & 31.4 & \textbf{29.2} & 37.0 & \textbf{5.5} & 0.26 & \textbf{0.02} & 0.08 & \textbf{0.03} & \textbf{15.69} & 15.85 \\
			\cline{2-16}
			\multicolumn{1}{|c|}{} & 2 & 0.14 & \textbf{0.32} & 7.0 & \textbf{1.3} & 34.3 & \textbf{29.8} & 28.8 & \textbf{4.7} & 0.17 & \textbf{0.02} & 0.13 & \textbf{0.05} & \textbf{15.03} & 15.16 \\
			\cline{2-16}
			\multicolumn{1}{|c|}{} & 3 & 0.28 & \textbf{0.34} & 7.2 & \textbf{0.0} & 32.5 & \textbf{30.2} & 32.7 & \textbf{4.8} & 0.29 & \textbf{0.02} & 0.09 & \textbf{0.02} & 16.15 & \textbf{16.14} \\
			\cline{2-16}
			\multicolumn{1}{|c|}{} & 4 & \textbf{0.22} & 0.20 & 8.5 & \textbf{6.0} & \textbf{32.3} & 36.6 & 32.1 & \textbf{5.6} & 0.23 & \textbf{0.07} & 0.11 & \textbf{0.08} & \textbf{15.78} & 18.37 \\
			\cline{2-16}
			\multicolumn{1}{|c|}{} & 5 & 0.22 & \textbf{0.30} & 3.2 & \textbf{1.1} & 40.1 & \textbf{37.2} & 36.5 & \textbf{5.1} & 0.18 & \textbf{0.01} & 0.07 & \textbf{0.02} & \textbf{20.41} & 20.48 \\
			\cline{1-16}
			\multicolumn{1}{|c|}{ \multirow{5}*{\textbf{Scenario (b)}} } & 1 & \textbf{0.32} & 0.18 & \textbf{0.3} & 15.3 & \textbf{46.8} & 61.6 & 31.2 & \textbf{6.2} & 0.20 & \textbf{0.11} & \textbf{0.05} & 0.11 & \textbf{23.50} & 28.52 \\
			\cline{2-16}
			\multicolumn{1}{|c|}{} & 2 & 0.20 & \textbf{0.32} & 5.7 & \textbf{0.9} & 56.8 & \textbf{44.8} & 32.9 & \textbf{4.9} & 1.73 & \textbf{0.01} & 0.06 & \textbf{0.03} & 25.14 & \textbf{24.56} \\
			\cline{2-16}
			\multicolumn{1}{|c|}{} & 3 & 0.28 & \textbf{0.29} & 0.7 & \textbf{0.6} & 63.1 & \textbf{61.0} & 34.1 & \textbf{5.3} & 3.14 & \textbf{0.02} & 0.05 & \textbf{0.02} & \textbf{33.28} & 33.50 \\
			\cline{2-16}
			\multicolumn{1}{|c|}{} & 4 & \textbf{0.41} & 0.32 & \textbf{0.0} & \textbf{0.0} & 59.9 & \textbf{56.6} & 35.7 & \textbf{4.9} & 2.82 & \textbf{0.01} & 0.06 & \textbf{0.03} & \textbf{30.39} & 30.69 \\
			\cline{2-16}
			\multicolumn{1}{|c|}{} & 5 & 0.28 & \textbf{0.32} & 0.7 & \textbf{0.6} & 76.9 & \textbf{66.2} & 37.7 & \textbf{5.4} & 1.84 & \textbf{0.02} & 0.05 & \textbf{0.02} & 37.86 & \textbf{35.61} \\
			\cline{1-16}
			\multicolumn{1}{|c|}{ \multirow{1}*{\textbf{Scenario (h)}} } & 1 & \textbf{0.33} & 0.32 & 1.8 & \textbf{1.1} & 22.1 & \textbf{18.8} & 58.3 & \textbf{5.0} & 0.16 & \textbf{0.01} & 0.09 & \textbf{0.05} & \textbf{9.85} & 10.13 \\
			\hline
	\end{tabular}}
\end{table*}

\subsection{Evaluation}
To reduce the randomness of evaluation, we select multiple sets of different start and goal poses to test local planners in each scenario except for the scenario depicted in Fig. \ref{datasets}(g), since this is a one-way environment. The configuration of the start and goal poses in each test is available on the benchmark website \cite{wen2020mrpb}. Tables \ref{statictable}, \ref{unknowntable}, and \ref{dynamictable} enumerate the quantitative statistics of local planning results in the static, partially unknown, and dynamic scenarios respectively. Scenarios (a)-(g) correspond to the scenarios shown in Fig. \ref{datasets}, and Scenario (h) corresponds to the dynamic scenario depicted in Fig. \ref{dynamic}(c)-(d). We use ``-'' to indicate the situation that the robot fails to reach the goal. \emph{It should be emphasized that the computation efficiency is related to the performance of the computing platform}. Therefore, the measurement results of this metric will be different on different machines.

\subsubsection{Comparison on Efficiency}
According to the evaluation results, it is concluded that TEB achieves superior performance than DWA in computational efficiency and motion efficiency. DWA needs to forward simulate and evaluate each pair of sampled velocities and does not take into account the environmental information during sampling. Therefore, considerable time is wasted in generating infeasible trajectories. While TEB takes the global path as the initial guess and obtains the local optimized trajectory through several iterations. As a result, planning with TEB is $ 8.94 $ times faster than planning with DWA. Furthermore, time optimality is explicitly considered in the optimization objectives of TEB. Therefore, TEB provides more efficient motion commands for the robot. Compared with DWA, the motion efficiency of TEB is increased by $ 9.2\% $ on average. 

\subsubsection{Comparison on Safety}
TEB employs a set of configurations to form a virtual band. The trajectory optimization is performed by applying artificial forces to the band, wherein the repulsive force stretches the band to avoid collision with obstacles. Therefore, the optimized path usually has a certain distance to obstacles. As shown in Tables \ref{statictable}-\ref{dynamictable}, TEB achieves better security performance in most cases. Because of the better clearance from obstacles, the total travel distance of TEB is relatively longer than that of DWA.

\subsubsection{Comparison on Flexibility}
The evaluation results indicate that TEB performs better flexibility than DWA. Especially in scenarios like mazes that require robots to turn continuously, the shortcomings of DWA are obvious. As mentioned before, DWA performs forward simulation by applying each pair of sampled velocities for some short time. When the robot navigates in a narrow space, it is possible that all sampling velocities are infeasible. As a result, the robot may fall into oscillation and fail to complete the navigation task, as shown in Table \ref{statictable}. For this problem, adaptively adjusting the forward simulation step size according to the complexity of environments and the density of obstacles is a promising solution. In addition, the performance of DWA is closely related to the weight of each item in the evaluation function. Currently, these weights are hand tuned by operators. A promising approach is to apply machine learning techniques such as Deep Q-Learning (DQN) \cite{mnih2013playing} to discover the proper mapping from action and environment to cost. Certainly, this kind of approach is self-supervised and requires a wealth of effective feature inputs and representative training data.

In summary, TEB performs better than DWA in terms of efficiency, safety, and flexibility. Such type of optimization-based local planners may be a better choice for indoor navigation than sampling-based local planners.

\section{Conclusion}
\label{conclusion}
In this paper, we newly propose a mobile robot local planning benchmark called MRPB 1.0 to evaluate mobile robot local planning approaches in a unified and comprehensive way. Various simulation scenarios are elaborately designed and three types of principled evaluation metrics are proposed. We present the application of the proposed benchmark in two local planners to show the practicality of the benchmark.


\bibliographystyle{IEEEtran}
\bibliography{ref}

\end{document}